\pdfoutput=1

\documentclass[letterpaper, 10 pt, conference]{ieeeconf}  

\IEEEoverridecommandlockouts                              

\usepackage{graphicx} 
\usepackage{amsmath} 
\usepackage{amsfonts}
\usepackage{autobreak}
\usepackage{algorithm}
\usepackage{algpseudocode}
\usepackage{lmodern}
\usepackage{cite}
\usepackage{makecell}
\usepackage{booktabs}

\title{\LARGE \bf
Exploration-efficient Deep Reinforcement Learning with Demonstration Guidance for Robot Control
}

\author{Ke Lin$^{1}$, Liang Gong$^{1}$, Xudong Li$^{1}$, Te Sun$^{1}$, Binhao Chen$^{1}$, Chengliang Liu$^{1}$,  \\Zhengfeng Zhang$^{2}$, Jian Pu$^{2}$, Junping Zhang$^{2}$ 
\thanks{$^{1}$Shanghai Jiao Tong University, Shanghai 200240, China. }%
\thanks{$^{2}$Fudan University, Shanghai 200433, China. }
}

\begin{document}
\maketitle
\thispagestyle{empty}
\pagestyle{empty}

\bibliographystyle{IEEEtran.bst}  

\begin{abstract}

Although deep reinforcement learning (DRL) algorithms have made important achievements in many control tasks, they still suffer from the problems of sample inefficiency and unstable training process, which are usually caused by sparse rewards. Recently, some reinforcement learning from demonstration (RLfD) methods have shown to be promising in overcoming these problems. However, they usually require considerable demonstrations. In order to tackle these challenges, on the basis of the SAC algorithm we propose a sample efficient DRL-EG (DRL with efficient guidance) algorithm, in which a discriminator $D(s)$ and a guider $G(s)$ are modeled by a small number of expert demonstrations. The discriminator will determine the appropriate guidance states and the guider will guide agents to better exploration in the training phase. Empirical evaluation results from several continuous control tasks verify the effectiveness and performance improvements of our method over other RL and RLfD counterparts. Experiments results also show that DRL-EG can help the agent to escape from a local optimum.

\end{abstract}
\section{INTRODUCTION}

Reinforcement learning (RL) \cite{intro:Sutton1998Reinforcement} enables intelligent agents to make good decisions in environments and make it possible for a robot to learn specific skills by interacting with environments. However, in many robotic control tasks, the observation space and action space are usually continuous and high-dimensional, so traditional RL algorithms usually cannot cope with these tasks very well. Recently, deep reinforcement learning (DRL) algorithms have achieved significant successes in many continuous control tasks and difficult sequential decision-making problems \cite{intro:duan2016benchmarking}. \par
In general, the original DRL algorithms can be divided into 2 types, i.e., policy-based algorithms such as vanilla policy gradient \cite{intro:sutton1999policy} and value-based algorithms such as deep Q-learning \cite{intro:dqn}. In order to combine the advantages of these two types of algorithms, actor-critic style algorithms were devised by many researchers \cite{intro:actor-critic}, such as advantage actor-critic (A2C), asynchronous advantage actor-critic (A3C) \cite{intro:a3c} and actor-critic with experience replay (ACER) \cite{intro:acer}. On the other hand, with the aim of alleviating the unstable training process of policy networks, trust region policy optimization (TRPO) \cite{intro:trpo} was proposed by Schulman, in which KL-Divergence was employed to constrain the update of the policy. Subsequently, proximal policy optimization (PPO), a simpler version, was proposed by Schulman as well \cite{intro:ppo}. In addition, deep deterministic policy gradient (DDPG) \cite{intro:ddpg} can be thought of as an extension of DQN in continuous action space. Additionally, twin delayed DDPG (TD3), an improved version of DDPG \cite{intro:td3}, was put forward by Fujimoto to alleviate the problem of dramatically overestimating Q-values and make the training process more stable. Moreover, soft actor-critic (SAC) \cite{intro:sac} appeared almost simultaneously with TD3, which is a sort of actor-critic style RL algorithm, featured by DDPG-style critics and maximum entropy regularization. It is worth emphasizing that SAC has achieved the state-of-the-art results in many robotic continuous control tasks. \par
Despite DRL algorithms having achieved significant success, these methods still suffer from the problems of sample inefficiency, difficulty of escaping from a local optimum and unstable training process \cite{intro:a13}, where the sample inefficiency problem is particularly serious. This is because agents cannot be rewarded in a timely manner. In addition, in the early stages of training, the policy is a stochastic policy, which cannot make good actions, so it results in a poor reward. Thus, if we can guide agents to take appropriate actions and guide agents to go to high-reward areas, the sample efficiency will be greatly enhanced. Based on this idea, there has been a sustained research activity in the combination of RL algorithms with expert demonstrations \cite{intro:a14,intro:a15,intro:a16,intro:a17,intro:a18}. Recently, it is common practice for DRL algorithms to put demonstration data into replay buffers {\cite{intro:a19,intro:a20,intro:a21}} or pre-train the policy network \cite{intro:a22,intro:a23}. However, these sorts of methods cannot make full use of demonstration data, limited by only using them in a supervised learning manner. To address this deficiency, absorbing some useful ideas from imitation learning \cite{intro:a24,intro:a25,intro:a26,intro:a27,intro:a28}, more advanced RLfD approaches \cite{intro:a17,intro:a18} were proposed. Many of them used reward shaping techniques to guide agents to explore high-reward areas, this is because rewards provide the most informative information about an environment \cite{intro:a29}. However, these methods usually require a lot of expert demonstrations. But for many robotic tasks, it is difficult to collect a large amount of demonstration data. Also, traditional robot control methods usually require complex dynamics modeling and sophisticated control algorithm design.\par
Thus, in this paper, we proposed a sample efficient DRL-EG (DRL with efficent guidance) algorithm , which can be used in robotic continuous control tasks. Our method, built on the SAC, is a novel method that can make effective use of demonstration data in agents’ exploration phase. Firstly, a discriminator $D(s)$ and a guider $G(s)$ will be built by demonstration data. Then, in the training phase, the discriminator $D(s)$ will judge whether the guider can give a good action in the current state $s$. If so, the agent will take the action output by the guider $G(s)$, otherwise, the agent will take an action by the policy. Finally, better policy can be achieved through better exploration. Empirical evaluation results on continuous control tasks verify the effectiveness and performance improvement of our method over other counterparts, such as SAC \cite{intro:sac}, PPO \cite{intro:ppo}, PPO with pre-training and DDPGfD \cite{intro:ddpg}. Experiments results also showed that DRL-EG method can help the agent escape from a local optimum.\par
The major contributions of this paper could be summarized as follows:
\begin{itemize}
	\item	Two types of discriminator $D(s)$ and guider $G(s)$, including neural network style and functional style, are proposed to guide the agent to better exploration in the environment.
	\item	We developed a sample efficient DRL algorithm named DRL-EG to tackle the problems in pure DRL algorithms, such as sample inefficiency and difficulty of escaping from a local optimum.
	\item	With the guided mechanism, DRL-EG achieves consistent
	improvement over other RL and RLfD counterparts on several continuous robotic control benchmarks.
\end{itemize} \par
The rest of paper is organized as follows: In Section 2, we provide the subject of maximum entropy RL framework. Then our proposed method will be detailed in Section 3. Empirical evaluations results and discussion will be demonstrated in Section 4. Finally,  conclusion will be represented in Section 5.

\section{PRELIMINARIES}

\subsection{Markov Decision Process (MDP)}
The proposed method is based on the framework of the infinite-horizon markov decision process (MDP), represented by the tuple $(S,A,P,r,\gamma)$, in which $S$ and $A$ represent state space and action space, respectively. The state transition probability $P(s'|s,a):S \times A \times S \rightarrow R$ denotes the probability density of transition to next state $s'$ after performing action $a$ in state $s$. $r$ is a reward function which maps state-action pairs to real numbers, i.e., $r(s,a): S \times A \rightarrow R$. $\gamma \in{(0,1)}$ is discount factor. \par
Usually, a stochastic policy $\pi (a|s):S \times A \rightarrow [0,1]$ for markov decision process (MDP) gives the probability of taking an action when in state $s$. $\rho(s):S \rightarrow R$ gives the probability density of the initial state $s$. An episode is a trajectory of an agent’s interaction with the environment, so an episode of a MDP can be defined by a sequence $\tau :\{s_0,a_0,r_0,s_1,a_1,r_1,...\}$. MDP used in RL is aimed at maximizing average returns:
\begin{equation}
	\pi^{*}=\arg \max_{\pi} \mathbb{E}_{\pi}\left[\sum_{t=0}^{\infty}{\gamma^tr\left(s_t,a_t\right)}\right] 
\end{equation} \par
In reinforcement learning, state value function $V^\pi (s)$ and state-action value function $Q^\pi (s,a)$ are introduced to evaluate the performance of a policy. $V^\pi (s)$ and $Q^\pi (s,a)$ can be given by the expression:
\begin{equation}
	V^\pi\left(s\right)=\mathbb{E}_{\pi}\left[\left.\sum_{t=0}^{\infty}{\gamma^tr(s_t,a_t)}\right|s_0=s\right]
\end{equation}
\begin{equation}
	Q^\pi\left(s,a\right)=\mathbb{E}_{\pi}\left[\left.\sum_{t=0}^{\infty}{\gamma^tr(s_t,a_t)}\right|s_0=s,a_0=a\right]
\end{equation}
where $\mathbb{E}_{\pi}$ means the expectation is taken with respect to the policy $\pi$. According to the definition of the state value function and the state-action value function, $Q^\pi(s,a)$ and $V^\pi(s)$ can be transformed from each other:
\begin{equation}
	Q^\pi\left(s,a\right)=r\left(s,a\right)+\gamma\mathbb{E}_{s' \sim P(\cdot | s,a)}\left[V^\pi(s')\right]
\end{equation}
\begin{equation}
	V^\pi\left(s\right)=\mathbb{E}_{a \sim \pi(\cdot|s)}\left[Q^\pi\left(s,a\right)\right]
\end{equation}
\subsection{Soft Policy Iteration}
Soft actor-critic algorithm is a sort of maximum entropy reinforcement learning algorithm \cite{intro:sac}, which has achieved the state-of-the-art performance in many continuous control tasks \cite{pre:sacApp}. Entropy is often used to indicate the degree of uncertainty of a distribution \cite{pre:entropy}, which can be represented by $\mathcal{H}\left(P\right)=\mathbb{E}_{x \sim P}\left[-\log{P(x)}\right]$. In maximum entropy reinforcement learning, the entropy term is added to the reward of each step, i.e., $r\left(s,a\right)+\alpha\mathcal{H}\left(\pi(\cdot|s)\right)$, which helps to improve the stability and robustness of the model.  \par
In this case, the objective is to maximize cumulative rewards and entropy of the policy: 
\begin{small}
	\begin{equation}
		\pi_{soft}^*=\arg{\max_\pi{\mathbb{E}_{\pi}\left[\sum_{t=0}^{\infty}{\gamma^t\left[r\left(s_t,a_t\right)+\alpha\mathcal{H}\left(\pi(\cdot|s_t)\right)\right]}\right]}}
	\end{equation}
\end{small}
Moreover, equation (4) and (5) can be modified to:
\begin{equation}
	Q_{soft}^\pi\left(s,a\right)=r\left(s,a\right)+\gamma \mathbb{E}_{s^\prime \sim P(\cdot|s,a)}\left[V^\pi(s')\right]
\end{equation}
\begin{equation}
	V_{soft}^\pi\left(s\right) = \mathbb{E}_{a \sim \pi(\cdot|s)}\left[Q^\pi\left(s,a\right)-\alpha\log{\pi(a|s)}\right]
\end{equation} \par
For a fixed policy $\pi$, calling (7) and (8) repeatedly can obtain $Q_{soft}^\pi\left(s,a\right)$. This process is called Q-iteration.
\section{METHODOLOGY}
\subsection{Demonstrations Guided Exploration Framework}
When people start to learn a new skill, such as swimming or shooting a basketball, people usually will not take actions blindly. Instead, they will imitate the actions of experts, and then explore some more suitable actions for them. Therefore, it is sensible to think that for RL algorithms, it is necessary to use demonstration data to guide the exploration of agents at the beginning of training, which can help agents learn faster. \par
Inspired by this, we proposed a demonstration data guided mechanism, which makes use of demonstration data to guide agents' actions in the training phase, as shown in Fig. 1. In state $s_t$, the discriminator $D(s)$ will judge whether the guider can give a good action in this state. If so, the agent will take the action output by the guider $G(s)$, otherwise, the agent will take an action output by the policy $\pi_\theta(a|s)$. \par 
Generally, the guider can be a neural network or a function, modeled using demonstration data, i.e., a series of state-action pairs $\{s_i^d,a_i^d\}$. The discriminator can be a distribution with respect to states or a function, modeled using a set of states $\{s_i^d\}$. This mechanism is mainly used when the average rewards of demonstration data far exceed the performance of the agent, especially in the early stage of training. With this mechanism, the sample efficiency will be enhanced, and the agent may be able to jump out of the local optimum. \par

Take an extreme example, suppose in a robot grasping task the reward is defined as: the reward is 1 when the end effector catches the target, otherwise the reward is 0. At the beginning stage of training, the policy is stochastic, so the agent can hardly approach the target. In this case, it is impossible for the agent to learn anything useful from rewards, since the rewards are all zero. Thus, if the guider can guide the robot to approach the target, the rewards will contain more useful information and the training speed will be greatly enhanced. \par

\subsection{Discriminator and Guider in Practice} 
\begin{figure}[t]
	\centering
	\framebox{\includegraphics[scale=0.45]{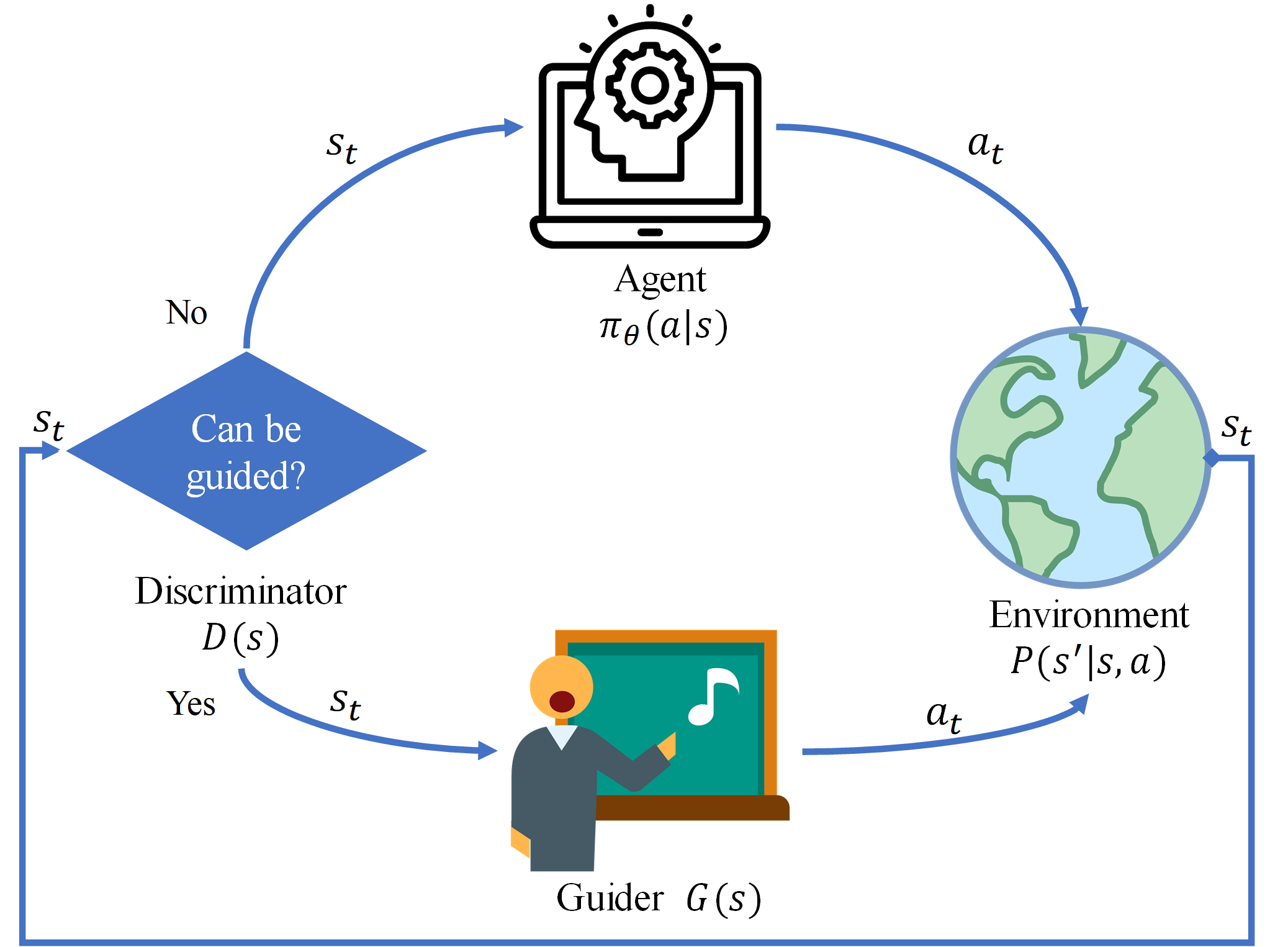}}
	\label{figframework}
	\caption{Demonstrations guided exploration framework.}
\end{figure}
Two types of discriminator are proposed, i.e., a state distribution and a function. For the distribution, it is modeled by a set of states $\left\{s_i^d\right\}$ in demonstration data, using gaussian mixture model (GMM) \cite{meth:GMM}:
\begin{equation}
P\left(s|\mu,\sigma^2\right)=\sum_{k=1}^{K}{\omega_k\phi(s|\mu_k,\sigma_k^2)}
\end{equation}
where $\phi\left(s\middle|\mu_k,\sigma_k^2\right) \sim \mathcal{N}(\mu_k,\sigma_k^2)$. Expectation maximization algorithm (EM algorithm) can be used to build a GMM. When the probability density of a state $s$ is greater than a threshold $T$, we can concluded that this state $s$ can be guided, otherwise it cannot be guided by the guider $G(s)$. It is worth noting that this kind of method is only applicable to the case when there are a huge number of state-action pairs. \par
For the case of relatively less demonstration data, such as only 1000 state-action pairs, we can use a functional discriminator:

\begin{equation}
D\left(s\right)=
\begin{cases}
1, & \min_{i}\|s-s_i^d\|_2 <= T_2 \\
0, & \min_{i}\|s-s_i^d\|_2 > T_2
\end{cases}
\end{equation}
where $T_2$ is a hyperparameter. For functional discriminator, we need to iterate through $\left\{s_i^d\right\}$ in demonstration data and calculate the Euclidean distance between these states $\left\{s_i^d\right\}$ and the state $s$. When the Euclidean distance between the state $s$ and state $\hat{s}$ in $\left\{s_i^d\right\}$ is short, we can assume that the state $s$ is similar to state $\hat{s}$. Thus, there is a similar state in demonstration data, and we can use the guider to guide the agent under state $s$. \par
Same as before, two types of guider are proposed, i.e., a neural network and a function. For the neural network, the input is a state and the output is an action, which is trained by state-action pairs $\left\{s_i^d,a_i^d\right\}$ in demonstration data. In the case of a small amount of demonstration data, a function is more practical, which can be given by:
\begin{equation}
G\left(s\right)=a_{i^*}^d+\xi,\ \ i^*=\arg \min_i{\|s-s_i^d\|_2}
\end{equation}
where $\xi \sim \mathcal{N}(0,\ \sigma^2)$. In this paper, we use functional $G(s)$ and $D(s)$, since we have collected no more than 1000 state-action pairs in each environment.


\subsection{DRL-EG Algorithm}
Due to the fact that the SAC algorithm has achieved the state-of-the-art performance in many continuous control tasks, we apply the demonstration data guided exploration mechanism to the SAC algorithm in order to get better performance, which leads to the proposed DRL-EG (DRL with efficient guidance) algorithm. \par
To simplify notation going forward, we will drop the mark \textit{soft} in $Q_{soft}^\pi\left(s,a\right)$ and $V_{soft}^\pi\left(s\right)$. Thus, $Q^\pi\left(s,a\right)$ and $V^\pi\left(s\right)$ mean the soft version of value functions. For continuous state space and action space, we usually use a neural network to approximate the tabular value function, so they can be denoted by $Q_\phi^\pi(s,a)$ and $V_\psi^\pi(s)$, where $\phi$ and $\psi$ represent the parameters to be optimized in neural networks. Similarly, a stochastic policy can also be represented by a neural network with independent Gaussian noise, denoted by $\pi_\theta(a|s)$. For the convenience of backpropagating the gradient, we denote the action output by the policy $\pi_\theta$ as:
\begin{equation}
f_\theta\left(s\right)=\pi_\theta^\mu\left(s\right)+\pi_\theta^\sigma\left(s\right)\odot\xi,\ \ \xi\sim \mathcal{N}(0,I)
\end{equation} 
which means the policy network parameterized by $\theta$ with independent noise can simultaneously output mean and standard deviation of action $a$ under state $s$. In order to maximize the expected rewards plus entropy, we can directly maximize $V_\psi^\pi(s)$ for all $s$ which is more practical, so we can rewrite the objective function as:
\begin{equation}  
J_\pi\left(\theta\right)=\mathbb{E}_{\pi}\left[Q_\phi^\pi\left(s,f_\theta\left(s\right)\right)-\alpha\log{\pi_\theta(f_\theta\left(s\right)|s)}\right]      
\end{equation}

When we get a mini batch of training data, the approximate gradient of (13) can be obtained by:
\begin{equation}
\begin{split}
\mathrm{\nabla}_\theta J_\pi(\theta)&=\frac{1}{N}\sum_{i=1}^{N}[\mathrm{\nabla}_{f_\theta(s_i)}Q_\phi^\pi(s_i,f_\theta(s_i))\mathrm{\nabla}_\theta f_\theta(s_i) \\
&-\alpha[\mathrm{\nabla}_\theta\log{\pi_\theta(a_i|s_i)} \\
&+\mathrm{\nabla}_{f_\theta(s_i)}\log{\pi_\theta(f_\theta(s_i)|s_i)}\mathrm{\nabla}_\theta f_\theta(s_i)]]
\end{split}
\end{equation} \par
For the evaluation of $Q_\phi^\pi(s,a)$ and $V_\psi^\pi(s)$, mean-square error (MSE) loss can be used to approach the target with a mini batch of samples. For Q-value, the target can be calculated by $y_q\left(s,a\right)=\ r\left(s,a\right)+\gamma V_\psi^\pi(s')$, and for V-value, the target is $y_v\left(s\right)=Q_\phi^\pi\left(s,f_\theta\left(s\right)\right)-\alpha\log{\pi_\theta\left(f_\theta\left(s\right)|s\right)}$. Thus, the loss function can be given by:
\begin{equation}
\begin{aligned}
&J_Q(\phi)=\mathbb{E}_{\pi}\left[\frac{1}{2}\left(Q_\phi^\pi\left(s,a\right)-y_q\left(s,a\right)\right)^2\right] \\
&J_V(\psi)=\mathbb{E}_{\pi}\left[\frac{1}{2}\left(V_\psi^\pi\left(s\right)-y_v(s)\right)^2\right]
\end{aligned}
\end{equation}
which can be optimized with stochastic gradient decent with a mini batch of samples:
\begin{equation}
	\begin{aligned}
		\nabla_\phi J_Q(\phi)&=\frac{1}{N}\sum_{i=1}^{N}[\nabla_\phi Q_\phi^\pi(s_i,a_i)\ast(Q_\phi^\pi(s_i,a_i) \\
		&-r(s_i,a_i)-\gamma V_\psi^\pi(s_i^\prime))] 
	\end{aligned} 
\end{equation}
\begin{equation}
	\begin{aligned}
		\nabla_\psi J_V(\psi)&=\frac{1}{N}\sum_{i=1}^{N}[\nabla_\psi V_\psi^\pi(s_i)\ast(V_\psi^\pi(s_i) \\
		&-Q_\phi^\pi(s_i,f_\theta(s_i))+\alpha\log{\pi_\theta(f_\theta(s_i)|s_i)})] 
	\end{aligned} 
\end{equation} \par
Thus, the proposed method can be described in detail, as shown in Algorithm 1.
\begin{algorithm}[t]
	\caption{DRL-EG Algorithm} %
	\textbf{Input}: Initialize parameter vectors of 5 neural networks, $Q_{\phi1}$, $Q_{\phi2}$, $V_\psi$, $\pi_\theta$, $V_{\psi\prime}$. Get demonstration buffer $D_{d}$ and average rewards $R_{demo}$ in $D_{d}$. Initialize an empty buffer $D_{b}$. Assign $R_{\pi_\theta} \leftarrow 0$.\\
	\textbf{Procedure}:
	\begin{algorithmic}[1]
		\State {Assign target parameters: $\psi^\prime\gets\psi$.}
		\For {each epoch}
			\State{Observe state $s$.}
			\If {$R_{demo}>R_{\pi_\theta}$}
				\State{Calculate $D(s)$ using $D_{d}$.}
				\If {$D(s)==1$}
					\State{Obtain $a_{guide}$ by calculating $G\left(s\right)$.}
				\Else
					\State{\textbf{goto} line 12.}
				\EndIf
			\Else
				\State{Select an action $a$ using $f_\theta\left(s\right)$.}
			\EndIf
			\State{The agent takes the action and get $(s,a,r,s',d)$.}
			\If {$d$ is true}
				\State{Reset the environment.}
			\EndIf
			\State{Store $(s,a,r,s',d)$ or $(s,a_{guide},r,s',d)$ in $D_{b}$.}
			\If{it's time to update}
				\For{$k=1,2,...,K$}
					\State{Sample a batch $\{s_i,a_i,r_i,s_i',d_i\}$ from $D_{b}$.}
					\State{Calculate $y_v(s_i)$ and $y_q(s_i,a_i)$.}
					\State{Update $\phi 1$, $\phi 2$ with $\nabla_\phi J_Q\left(\phi\right)$.}
					\State{Update $\psi $ with $\nabla_\psi J_V\left(\psi\right)$.}
					\State{Update $\theta$ with $\mathrm{\nabla}_\theta J_\pi\left(\theta\right)$.}
					\State{Update $\psi'$ with $\psi^\prime\gets\rho\psi^\prime+\left(1-\rho\right)\psi$.}
				\EndFor
				\State {Calculate $R_{\pi_\theta}$ by testing $\pi$ in the Env.}
			\EndIf
		\EndFor
	\end{algorithmic}
	\textbf{Output}: {$\pi_\theta(s)$.}
\end{algorithm} \par
\section{EXPERIMENTS AND DISCUSSION}
\begin{figure*}[t]
	\centering
	\framebox{\includegraphics[scale=0.7]{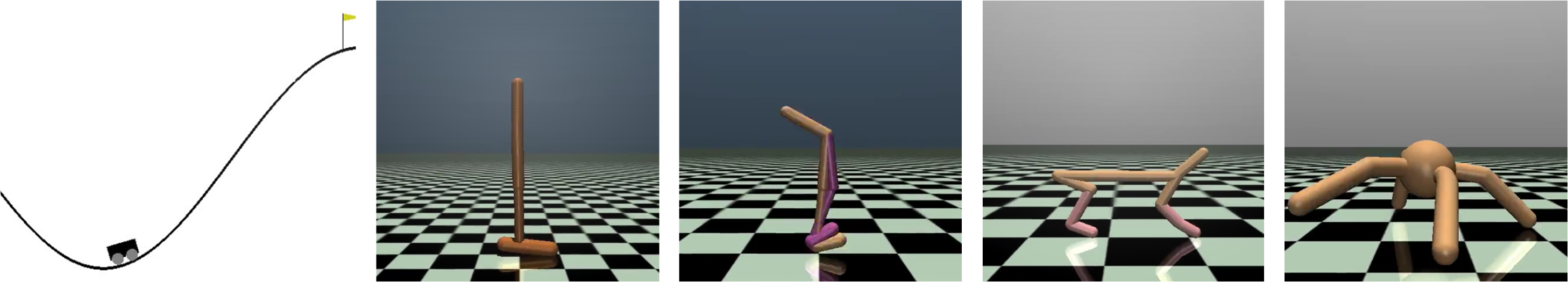}}
	\label{figenvs}
	\caption{Continuous control tasks environments in OpenAI gym used in this paper. MountainCarContinuous: the goal is to make the car reach the mountaintop. Hopper, Walker2d, HalfCheetah and Ant: the goal is to make the agent move as fast as possible.}
\end{figure*}
\begin{figure*}[t]
	\centering
	\framebox{\includegraphics[scale=0.35]{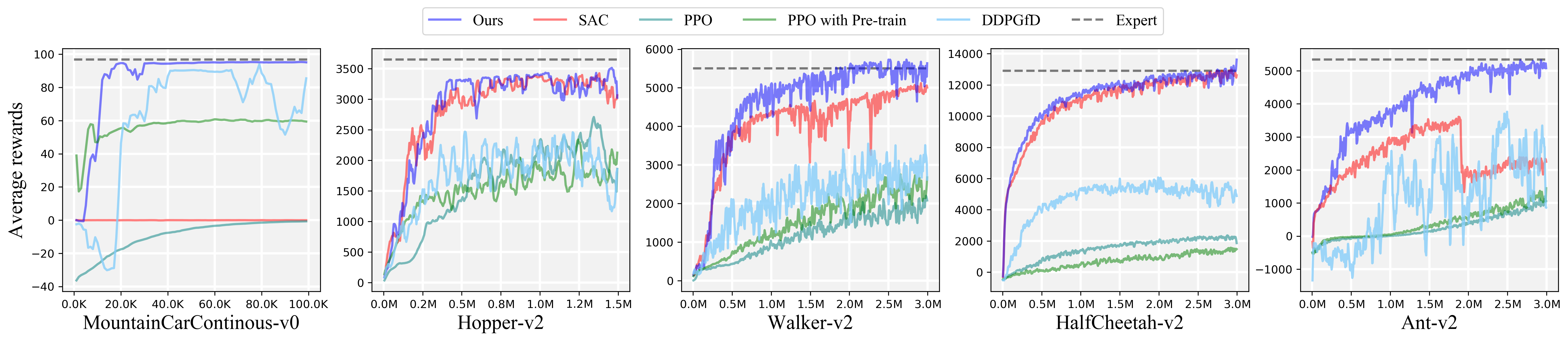}}
	\label{figexp1}
	\caption{Performance of our method versus the SAC and other RL and RLfD counterparts under continuous robotic control tasks. The abscissa indicates the number of steps, a step means one interaction with the environment. Each curve is averaged in three experiments.}
\end{figure*}

For the experiments below, we aim at investigating the following questions: 
\begin{enumerate}
	\item	Can the proposed algorithm achieve better performance than other RL and RLfD counterparts?
	\item	Under what circumstances does this mechanism work, and under what circumstances does it fail?
	\item	Can the guided exploration mechanism help the agent to jump out of the local optimum?
\end{enumerate}

\subsection{Experiments Configuration}
\begin{table}[b]
	\caption{Action and state space dimensions of environments}
	\begin{center}
		\begin{tabular}{ccc}
			\toprule
			\makecell[c]{Environment} & \makecell[c]{Observation \\ space dimension} & \makecell[c]{Action \\ space dimension} \\
			\midrule
			\makecell*[c]{MountainCarC-\\ontinuous-v0} & 2 & 1\\
			\makecell[c]{Hopper-v2} & 11 & 3\\
			\makecell[c]{Walker2d-v2} & 17 & 6\\
			\makecell[c]{HalfCheetah-v2} & 17 & 6\\
			\makecell[c]{Ant-v2} & 111 & 8\\
			\bottomrule
		\end{tabular}
	\end{center}
\end{table}
To answer the first two questions, we first carried out the proposed algorithm and the SAC to a series of continuous control tasks ranging from low-dimensional classical control tasks to challenging high-dimensional robotic control tasks in the OpenAI Gym \cite{openaigym}, some of which are implemented using MuJoCo, a physics engine aiming to facilitate research in robotics \cite{mujoco}. The description of these environments is shown in Fig. 2. More specifically, dimensions of the action space and state space of these environments are shown in Table 1. \par
Considering the randomness of the initialization of neural networks, for more persuasive evaluation and fair comparison, each experiment was implemented three times, so the curves are shown by taking the average of these three curves. Also, our method has the same experimental parameters as the SAC. We also compared our method with other RL and RLfD algorithms, such as proximal policy optimization (PPO) \cite{intro:ppo}, PPO with pre-train and deep deterministic policy gradient from demonstration (DDPGfD) \cite{intro:a20}. \par
In deep reinforcement learning, the training course is usually unstable \cite{meth:robust}. In order to dig deeper into the effect of the proposed mechanism on the training process and answer the third question, we also carried out the proposed algorithm and the SAC in the Ant-v2 environment under four fixed random seeds. In each comparative experiment, not only do they have identical training parameters, but also have the same initialization parameters of neural networks. The only difference between each comparative experiment is the demonstrations guided exploration mechanism. \par
In addition, the demonstration data is collected as follows:
\begin{itemize}
	\item	For MountainCarContinuous-v0 environment, the demonstration data was collected by manual marking. The action in demonstration data is defined as: when the velocity is less than 0, the demonstration action is -0.5, whereas the demonstration action is 0.5.
	\item	For environments built by Mujoco, the demonstration data was obtained by implementing the best trained SAC or TD3 (Twin Delayed DDPG) \cite{intro:td3} agent in environments. 
	\item	The demonstration data for all environments contains 1000 state-action pairs. 
\end{itemize}
\begin{figure*}[t]
	\centering
	\framebox{\includegraphics[scale=0.4]{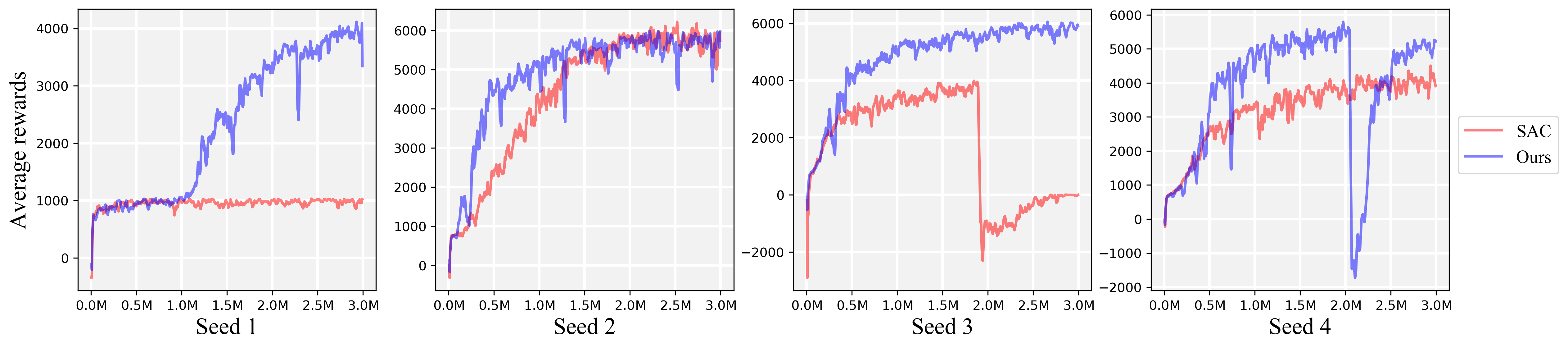}}
	\label{figexp2}	
	\caption{Performance of our method versus SAC under Ant-v2 environment with four different random seeds.}
\end{figure*}
\subsection{Comparative Evaluations}
Fig. 3 shows the performance of our method versus the SAC and other RL and RLfD counterparts under continuous control tasks. The performance is defined as the average cumulative rewards of several trajectories in the training course. \par
In the MountainCarContinous environment, the SAC and PPO can hardly learn anything useful from rewards. This is because the rewards given by the environment are so sparse. In this environment, positive rewards are only given when the car reaches the top of the mountain, and rewards will be reduced when the car takes an action, as a loss of the car’s movement. Therefore, as in the early stage of exploration the agent has never reached the top of the mountain and the rewards it receives have always been nonpositive, so the agent decides not to take any action in the environment to ensure maximum rewards. Thus, the rewards obtained by the SAC agent and PPO agent are 0 at the end of training. Fortunately, the proposed mechanism can guide the agent to go to high reward areas to a certain extent, so once the agent reaches the top of the mountain and gets high rewards, it can learn from those rewards and obtain a better policy. In addition, PPO with pre-train and DDPGfD also have the same guiding ability, but the final performance of these two methods are not as good as the proposed algorithm. \par
In the Hopper and HalfCheetah environments, the performance of the SAC is almost indistinguishable from the proposed algorithm. We think the reason for this case is that the SAC is already able to achieve good performance in these environments. For better performance the additional guided mechanism is not necessary. \par
In the Walker and Ant environments, our method outperforms all other RL and RLfD algorithms. Moreover, the proposed method can almost achieve the same performance as the demonstration data at the end of training. In addition, the SAC and DDPGfD have encountered the problem of unstable training course in the Ant environment, which is usually caused by the nonstationarity of the incoming data and the difficulty of obtaining steady improvement \cite{exp:robust}. Sometimes, small differences in parameter space of the policy will make large differences in performance. However, in these environment, the DRL-EG algorithm shows better training stability. \par
As a consequence, based on the results in Fig. 3, we can conclude that the proposed algorithm can get better performance than other counterparts. In addition, the DRL-EG algorithm can not only learn efficiently in a sparse reward environment such as in MountainCarContinuous, but also achieve more stable training stability. It is worth noting that when the SAC can achieve excellent performance in some environments, the guided exploration mechanism will not significantly improve agent’s performance, such as in Hopper and HalfCheetah environments. \par
Fig. 4 shows the performance of our method versus the SAC under Ant-v2 environment with four different random seeds. In each picture, the training parameters and experiment settings are totally identical, except for the fact that our method contains the guided exploration mechanism. For random seed 1, it is obvious that the SAC was trapped in a local optimum. In the early stage of training, our method was also trapped in a local optimum. However, due to the guided exploration mechanism, our method successfully jumped out of the local optimum. For random seed 2, our method converged faster than the SAC, and they had the same performance at the end of training. For random seed 3 and 4, both algorithms had experienced the unstable training problem, but our method achieved better performance at the end of training. \par
Therefore, based on the results in Fig. 4, we can make the observation that our method can help agents to escape from a local optimum and get better results. \par

\section{CONCLUSIONS}

In order to tackle the problems of low sample efficiency, easy to fall into a local optimum and training instability in pure RL algorithms, in this paper we propose a sample efficient DRL-EG algorithm, which can be used in robotic continuous control tasks. The main intuition is that expert demonstrations can be used to guide an agent to go to high reward areas, which helps to reduce ineffective exploration and improves sample efficiency. Therefore, we built a discriminator $D(s)$ and a guider $G(s)$ using demonstration data. Then, in the training phase, the discriminator $D(s)$ will judge whether the guider can give a good action in the current state $s$. The guider $G(s)$ will guide agents to take good actions. Finally, better policy can be achieved through better exploration. The experimental results on several continuous control tasks show that the DRL-EG can achieve better performance than other RL and RLfD counterparts and can help pure DRL algorithms escape from a local optimum. Further exploration on combining our work with computer vision to enable a real robot to grasp a target in dynamic open scenes or making a robot move in a real environment could be a new direction of the future work. \par

\section*{ACKNOWLEDGMENT}
We would like to thank Yanrui Jin and Kai Zhu for insightful discussions. We would also like to thank Duantengchuan Li and Christopher Wu for their valuable work of sample collection.


\addtolength{\textheight}{-4cm}   

\bibliography{root.bib}  
\end{document}